\documentclass[dvipsnames,format=sigconf,anonymous=false,review=false]{acmart} 

\def\vec#1{\mathchoice{\mbox{\boldmath$\displaystyle#1$}}
{\mbox{\boldmath$\textstyle#1$}}
{\mbox{\boldmath$\scriptstyle#1$}}
{\mbox{\boldmath$\scriptscriptstyle#1$}}}

\usepackage{booktabs} 
\usepackage{amsmath}
\usepackage{algorithm}
\usepackage[noend]{algpseudocode}
\usepackage[T1]{fontenc}
\usepackage{tabularx}
\usepackage{multirow}
\usepackage{array}
\usepackage{comment}
\usepackage{pbox}

\usepackage{color,soul}
\usepackage{nameref}

\usepackage{pgfplots}
\usepackage{graphicx}
\usepackage{subcaption}
\usepackage{tikz}

\usepackage{url}
\usepackage{bm}

\usepackage{comment}

\definecolor{darkgreen}{rgb}{0.30, 0.50, 0.0}

\makeatletter
\renewcommand{\ALG@name}{Pseudocode}
\makeatother

\newcolumntype{L}[1]{>{\raggedright\let\newline\\\arraybackslash\hspace{0pt}}m{#1}}
\newcolumntype{C}[1]{>{\centering\let\newline\\\arraybackslash\hspace{0pt}}m{#1}}
\newcolumntype{R}[1]{>{\raggedleft\let\newline\\\arraybackslash\hspace{0pt}}m{#1}}

\graphicspath{{images/}}

\AtBeginDocument{%
  \providecommand\BibTeX{{%
    \normalfont B\kern-0.5em{\scshape i\kern-0.25em b}\kern-0.8em\TeX}}}

\copyrightyear{2025}
\acmYear{2025}
\setcopyright{cc}
\setcctype{by}
\acmConference[GECCO '25]{Genetic and Evolutionary Computation
Conference}{July 14--18, 2025}{Malaga, Spain}
\acmBooktitle{Genetic and Evolutionary Computation Conference (GECCO '25),
July 14--18, 2025, Malaga, Spain}\acmDOI{10.1145/3712256.3726428}
\acmISBN{979-8-4007-1465-8/2025/07}

\hypersetup{draft}
\begin{document}
\title[Moving between high-quality optima using multi-satisfiability characteristics in hard-to-solve Max3Sat instances]{Moving between high-quality optima using multi-satisfiability characteristics in hard-to-solve Max3Sat instances}


   \author{Jedrzej Piatek}
 	    \affiliation{
    		\institution{Wroclaw Univ. of Science and Techn.}
    		\city{Wroclaw}
    		\country{Poland} 
	   }
 	\email{jedrzej.piatek1@gmail.com}

\author{Michal W. Przewozniczek}
\affiliation{
		\institution{Wroclaw Univ. of Science and Techn.}
		\city{Wroclaw}
		\country{Poland} 
	}
	\email{michal.przewozniczek@pwr.edu.pl}

\author{Francisco Chicano}
        \affiliation{
          \institution{University of M\'alaga}
          \city{Malaga} 
          \country{Spain} 
          \postcode{29071}
        }
        \email{chicano@uma.es}

\author{Renato Tin\'os}
        \affiliation{
          \institution{University of S\~ao Paulo}
          \city{Ribeir\~ao Preto} 
          \country{Brazil}
          \postcode{14040901}
        }
        \email{rtinos@ffclrp.usp.br}





	\begin{abstract}
        Gray-box optimization proposes effective and efficient optimizers of general use. To this end, it leverages information about variable dependencies and the subfunction-based problem representation. These approaches were already shown effective by enabling \textit{tunnelling} between local optima even if these moves require the modification of many dependent variables. Tunnelling is useful in solving the maximum satisfiability problem (MaxSat), which can be reformulated to Max3Sat. Since many real-world problems can be brought to solving the MaxSat/Max3Sat instances, it is important to solve them effectively and efficiently. Therefore, we focus on Max3Sat instances for which tunnelling fails to introduce improving moves between locally optimal high-quality solutions and the region of globally optimal solutions. We analyze the features of such instances on the ground of phase transitions. Based on these observations, we propose manipulating clause-satisfiability characteristics that allow connecting high-quality solutions distant in the solution space. We utilize multi-satisfiability characteristics in the optimizer built from typical gray-box mechanisms. The experimental study shows that the proposed optimizer can solve those Max3Sat instances that are out of the grasp of state-of-the-art gray-box optimizers. At the same time, it remains effective for instances that have already been successfully solved by gray-box.



	\end{abstract}

    \ccsdesc[500]{Computing methodologies~Artificial intelligence}
    \ccsdesc[500]{Computing methodologies~Discrete space search}
    \ccsdesc[500]{Computing methodologies~Randomized search}
    \ccsdesc[500]{Theory of computation~Evolutionary algorithms}
    \ccsdesc[500]{Mathematics of computing~Evolutionary algorithms}

	\keywords{Gray-box optimization, Genetic Algorithms, Iterated Local Search, Evolutionary Algorithms, Optimization, Phase Transitions }
	
	\maketitle
	
	\section{Introduction}
\label{sec:intr}

Gray-box optimization aims to propose optimizers that are highly effective, efficient and preserve the robustness of black-box optimizers. To this end, gray-box optimization exploits the knowledge about variable dependencies and the subfunction-based representation that is frequently available for many real-world problems and is commonly used by exact solvers \cite{whitleyNext}. Using such problem-specific information (but generalized in its form) makes gray-box optimizers effective, efficient, and robust.\par

Some gray-box optimizers focus on utilizing masks clustering dependent variables to improve the effectiveness of variation operators \cite{pxForBinary,ilsDLED}. Such masks allow finding the improving moves for high-quality solutions that would be hard to make otherwise. Frequently, they allow jumping from one locally optimal solution to the other of higher quality, which is denoted as \textit{tunnelling}. Tunnelling was shown to be particularly useful in Max3Sat optimization \cite{max3satPlateaus}. However, tunnelling has its limitations. The masks offered by state-of-the-art gray-box mechanisms are unable to tunnel between optima that are too far away from each other in terms of Hamming distance. Therefore, we analyze the features of hard-to-optimize Max3Sat instances using tools supported by phase transitions \cite{backbone,phaseGabriela}. We focus on relatively small instances (yet non-trivial and not toy-sized) that we have identified as hard to make our analyzis tractable. Using our observations, we formulate clause-satisfiability characteristics that reflect the features of globally optimal solutions for hard-to-solve instances. On this base, we propose the mechanism that allows connecting distant solution space regions containing high-quality solutions. Finally, we assemble our proposition with state-of-the-art gray-box mechanisms to propose the Max3Sat Optimizer With Clause-Satisfiability Manipulation (MOCSM). The experimental results (we use two significantly different Max3Sat instance generators) show that MOCSM can successfully solve those instances that are out of the grasp of the currently available state-of-the-art gray-box mechanisms. At the same time, the proposed mechanisms do not deteriorate MOCSM performance for those instances that have already been solved effectively.\par



As the additional side-effect input of this work, we propose estimating how hard it is to solve a given Max3Sat instance. The proposed procedure considers the same instance features as those that led to proposing clause-satisfiability characteristics. We confirm that the correlation between these features' strength and the median fitness function evaluation number (FFE) necessary to solve a given instance by the state-of-the-art black-box optimizer is statistically significant. Such results confirm that our intuitions and observations are valid. Additionally, we support a mechanism that may be beneficial for future works concerning benchmark formulation and optimizers evaluation.\par

The rest of this work is organized as follows. The next section presents related work. In Section \ref{sec:backbone}, we analyze the features of easy- and hard-to-solve Max3Sat instances. MOCSM and the procedure for the Max3Sat difficulty measurement are proposed in the fourth and fifth sections, respectively. Section \ref{sec:results} presents the experimental verification of our propositions. Finally, the last section concludes this work and proposes the most promising future work directions.

\section{Related Work}
\label{sec:relWork}
Gray-box optimization is particularly useful for handling \textit{k}-bounded pseudo-Boolean problems, which can be represented as a sum of subfunctions taking no more than \textit{k} arguments \cite{transTokBounded}. The \textit{additive form} is convenient to represent such problems:
    \begin{equation}
        \label{eq:additive}
        f(\vec{x})=\sum_{s=1}^{S} f_s(\vec{x}_{I_s}),
    \end{equation}
    where $\vec{x}=(x_1,x_2,\ldots,x_{n})$ is a binary vector of size $n$, $I_s$ are subsets of $\{1,...,n\}$, which do not have to be disjoint and $S$ is the number of these subsets.\par

To decompose \textit{k}-bounded pseudo-Boolean problems, \textit{Walsh decomposition} \cite{heckendorn2002} is useful. It allows defining any function as:

    \begin{equation}
    \small
    \label{eq:walsh-decomposition}
    f(\vec{x}) = \sum_{i=0}^{2^n-1} w_i \varphi_i(\vec{x}) 
    \end{equation}
    where $w_i \in \mathbb{R}$ is the $i$th Walsh coefficient, $\varphi_i(\mathbf{x}) =(-1)^{\mathbf{i}^\mathrm{T}\mathbf{x}}$ defines a sign, and $\mathbf{i} \in \{0,1\}^n$ is the binary representation of index $i$.  \par

\begin{table}  
    \caption{VIG for the Max3Sat instance example and the PX results for $\vec{x_a}=110101$, $\vec{x_b}=010000$, and $m_{a,b,2}=\{4,6\}$}
    \label{tab:vigPxExampl}
    \scriptsize
    \hspace{-3em}
    \begin{subtable}{.22\textwidth}
        \begin{tabular}{l|cccccc}
          & \textbf{1} & \textbf{2} & \textbf{3} & \textbf{4} & \textbf{5} & \textbf{6} \\
          \hline
        \textbf{1} & x & 1 & 1 & 0 & 0 & 0 \\
        \textbf{2} & 1 & x & 1 & 0 & 1 & 0 \\
        \textbf{3} & 1 & 1 & x & 0 & 1 & 0 \\
        \textbf{4} & 0 & 0 & 0 & x & 1 & 1 \\
        \textbf{5} & 0 & 1 & 1 & 1 & x & 1 \\
        \textbf{6} & 0 & 0 & 0 & 1 & 1 & x
        \end{tabular}
        \caption{Exemplary VIG}
        \label{tab:vigPxExampl:vig}
    \end{subtable}
    \hspace{-2em}
    \begin{subtable}{.22\textwidth}
    \centering
       \begin{tabular}{c|cccc}
            \textbf{clause} & $\vec{x_a}$ & $\vec{x_b}$ & $\vec{x_a}'$ & $\vec{x_b}'$ \\
            \hline
            $x_1 \lor \neg x_2 \lor x_3$    & 110  & 010  & 110 & 010 \\
            $\neg x_2 \lor x_3 \lor x_5$      & 100 & 100  & 100    & 100    \\
            $\neg x_4 \lor x_5 \lor \neg x_6$ & 101 & 000 & 000 &  101
    \end{tabular}
    \caption{Subfunction arguments}
    \label{tab:vigPxExampl:subf}
    \end{subtable}
\end{table}

Based on the results of Walsh decomposition, we consider variables $x_g$ and $x_h$ dependent if there exists a Walsh coefficient, which mask groups these two variables. Max3Sat is a sum of 3-element clauses where each element can be negated or not, e.g., $(x_1 \lor \neg x_2 \lor x_3) + (\neg x_2 \lor x_3 \lor x_5) + (\neg x_4 \lor x_5 \lor \neg x_6)$.
Thus, each clause can be considered as a subfunction in the additive form. Walsh decomposition will identify variables that are arguments of a given clause as dependent on each other. Such information can be stored in the Variable Interaction Graph (VIG) \cite{whitleyNext} that can be represented by a square matrix. Its entries equal '1' if a given variable pair is dependent or '0', otherwise. Table \ref{tab:vigPxExampl:vig} presents VIG for the above Max3Sat instance.\par

VIGs can be employed to support masks for variation operators. VIG-based masks are expected to cluster dependent variables because the sets of these variables may have a joint influence on the optimized function value. VIG-based perturbation (VIGbp) is proposed as a mechanism for producing perturbation masks in Iterated Local Search (ILS) \cite{ilsDLED}. In VIGbp, we randomly choose a single gene $x_r$, and we group it with all variables dependent on it concerning the VIG. If the size of such mask is larger than the user-defined threshold, then we randomly remove genes from this mask until the mask size is acceptable. For instance, concerning VIG presented in Table \ref{tab:vigPxExampl:vig} and $x_r=3$, the mask would be $\{1,2,3,5\}$. If the maximum mask size is three, then the mask for $x_r=3$ can be $\{1,3,5\}$. After perturbing the genotype, ILS optimizes it using local search, e.g., the First Improvement Hill Climber (FIHC) \cite{P3Original,FIHCwLL}. Its procedure is relatively simple. It flips variables in a random order. If flipping a variable yields better fitness, the modification is preserved or rejected otherwise. \par

Partition crossover (PX) \cite{pxForBinary} is a mixing operator utilizing VIG. To mix individuals $\vec{x_a}$ and $\vec{x_b}$, PX finds mixing masks in the following manner. All dependencies for which at least one variable is of equal value in both individuals, i.e., $\vec{x_a}(i)=\vec{x_b}(i)$, where $\vec{x}(i)$ denotes the value of the \textit{i}th variable of individual $\vec{x}$, are removed from VIG. Then, the connected components of the remaining VIG are used to define a recombination mask. Consider PX-based mask  $m_{a,b}$, for mixing $\vec{x_a}$ and $\vec{x_b}$. We can create two offspring individuals, $\vec{x_a}'$ and $\vec{x_b}'$, where $\vec{x_a}'$ is a copy of $\vec{x_a}$ with genes from $\vec{x_b}$ marked by $m_{a,b}$, which we can represent as $\vec{x_a}' = \vec{x_a} + m_{a,b}(\vec{x_b})$. $\vec{x_b}'$ is created analogously.\par

Using PX mixing masks we are guaranteed that the offspring individuals will have the following feature. The arguments of each subfunction $\vec{x}_{I_s}$ will be the same as in $\vec{x_a}$ or $\vec{x_b}$ \cite{dgga}. It is important for the optimization of \textit{overlapping problems} (the problems which subfunctions share their variables) because it allows mixing solutions without breaking the subfunction argument value sets. This feature can be also utilized to discover variable dependencies in black-box optimization \cite{dgga}. For the \textit{k}-bounded problems, if VIG contains dependencies arising from Walsh decomposition, then the fitness of individuals resulting from PX will satisfy the following relation: $f(\vec{x_a}) + f(\vec{x_b}) = f(\vec{x_a}') + f(\vec{x_b}')$.\par

Consider an example of PX for the Max3Sat instance presented above and individuals $\vec{x_a}=110101$ and $\vec{x_b}=010000$. After zeroing all rows and columns referring to variables 2, 3, and 5 that are equal in $\vec{x_a}$ and $\vec{x_b}$, we will obtain two possible masks $m_{a,b,1}=\{1\}$ and $m_{a,b,2}=\{4,6\}$. Table \ref{tab:vigPxExampl:subf} presents subfunction arguments for $\vec{x_a}$, $\vec{x_b}$, $\vec{x_a}'$ and $\vec{x_b}'$ if $m_{a,b,2}$ is used for mixing. Note that all argument value sets in $\vec{x_a}'$ and $\vec{x_b}'$ were inherited from parent individuals, and $f(\vec{x_a}) + f(\vec{x_b}) = f(\vec{x_a}') + f(\vec{x_b}') \rightarrow 1 + 1 = 2 + 0$.\par

Parameter-less Population Pyramid (P3) \cite{P3Original} proposes a new strategy of population management. In P3, population resembles a pyramid and each of its \textit{levels} is a separate subpopulation of individuals. In each iteration, a new individual (denoted as \textit{climber}) is created randomly, optimized by FIHC and added to the lowest pyramid level unless such an individual does not exist in the population yet. Then, the climber is mixed with the individuals on subsequent levels. If any of these operations improves individual, its improved copy is added one level higher (if such level does not exist, it is created). P3 is a black-box optimizer and to generate mixing masks it uses Statistical Linkage Learning \cite{ltga,dsmga2} that analyzes the occurence of gene value pairs in the population of individuals.

\section{Backbone-based analysis of hard- and easy-to-solve Max3Sat instances}
\label{sec:backbone}

Phase transitions \cite{phaseGabriela} aims at understanding why particular problem instances are hard- or easy-to-solve. The concept of \textit{backbone} is proposed in \cite{backbone}. To compute the backbone, we need to know some (preferably all) optimal solutions. Backbone is a common part of genotypes that refer to optimal solutions and can be interpreted as a region in solution space containing optimal solutions. Thus, the more genes a given solution has in common with the backbone, the closer it is (in terms of Hamming distance) to the region containing optimal solutions. Therefore, it can be considered valuable even if its fitness is low.\par

Backbone is a useful tool for the theoretical analyzis of Max3Sat instances because it requires knowledge concerning optimal solutions. Thus, it is not useful for practical use. Therefore, we aimed to identify those features of hard-to-solve Max3Sat instances that correlate with covering the backbone.\par

To generate optimal and near-optimal solutions for the considered instances, we used a gray-box optimizer denoted as \textit{ILS+P3+PX} (IPP). It integrates the ILS-like solution improvement procedure based on perturbation and local search with PX and the P3-like population management. Its detailed description is supported in Section \ref{sec:wmo:ipp}. To generate optimal solutions and the \textit{high-quality solution set}, IPP was executed for 1 hour in a single thread on Intel core i7-14700KF, 64GB RAM computer. Considered solutions were divided into two groups: solutions with only one unsatisfied clause and solutions with more than one unsatisfied clause but with at least 99.5\% of satisfied clauses. For instance, for 150-variable instances with 640 clauses (standard clause ratio $cr=4.27$) we considered solutions with at least 636 satisfied clauses.\par

For each solution in the high-quality solution set, we calculated how many of its gene values are common with the backbone. Additionally, we calculated the number of clauses that were satisfied by one, two, and three variables, respectively.


\subsection{Easy-to-solve instances}
\label{sec:backbone:easy}

\begin{figure}[h]
    \begin{subfigure}[b]{0.49\linewidth}            
        \centering
        \includegraphics[scale=0.30]{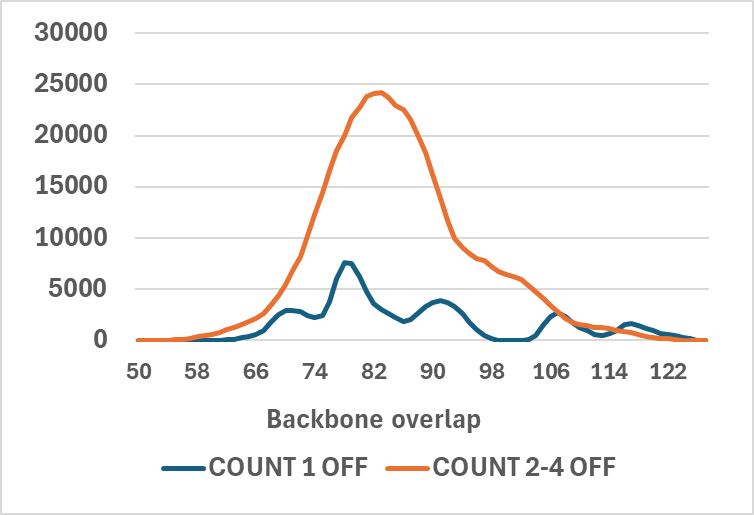}
        \caption{$Seed=0$}
        \label{fig:backbone:easy:file1}
    \end{subfigure}
    \begin{subfigure}[b]{0.49\linewidth}
        \centering
        \includegraphics[scale=0.30]{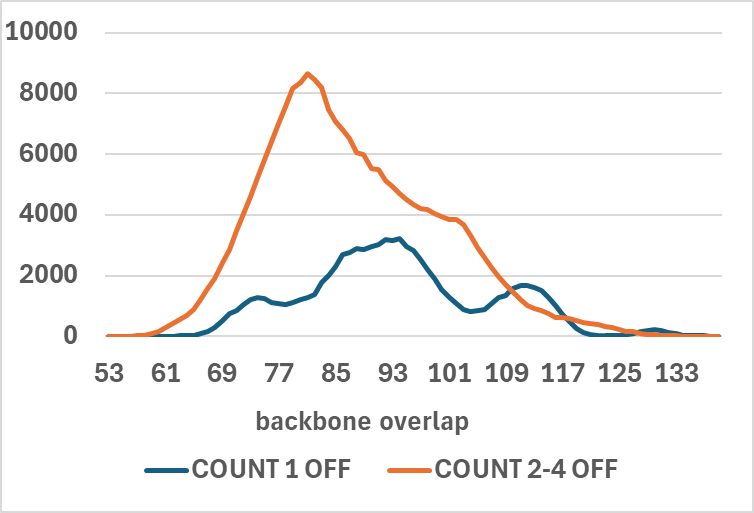}
        \caption{$Seed=1$}
        \label{fig:backbone:easy:file2}
    \end{subfigure}
    \caption{Backbone overlap distribution for the easy-to-solve instances (X axis: backbone overlap, Y axis: number of solutions) }
    \label{fig:backbone:easy:files}
\end{figure}

In Fig. \ref{fig:backbone:easy:files}, we present two backbone overlap distributions of the high-quality solution set. We consider solutions with one unsatisfied clause (the blue line) and more unsatisfied clauses (the orange line). Most of these solutions overlap with the backbone by approximately 85 variable values (the backbone size is 126 and 139 for Fig. \ref{fig:backbone:easy:file1} and Fig. \ref{fig:backbone:easy:file2}, respectively). However, for both distributions, there is a continuous string of solutions from the middle of the distribution to those that fully or almost fully cover the backbone. Thus, it seems intuitive that tunnelling between these local optima (e.g., using PX or the VIG-aware ILS) is possible because we can slide from one high-quality solution to the other without making large changes in the subsequent steps.\par

Note that the right side of the graphs has a very low number of solutions. The explanation of this phenomenon is as follows. If a solution fully or almost fully covers the backbone, then the nearest locally optimal solution is globally optimal. Thus, such solutions will not be considered in the presented graphs.\par

\subsection{Hard-to-solve instances}
\label{sec:backbone:hard}

\begin{figure*}[h]
    \begin{subfigure}[b]{0.33\linewidth}            
        \centering
        \includegraphics[scale=0.45]{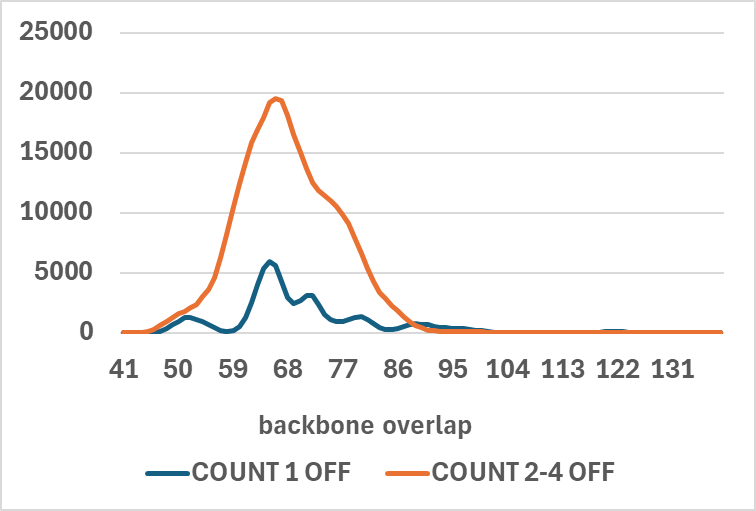}
        \caption{$Seed=2$}
        \label{fig:backbone:hard:file2}
    \end{subfigure}
    \begin{subfigure}[b]{0.33\linewidth}
        \centering
        \includegraphics[scale=0.45]{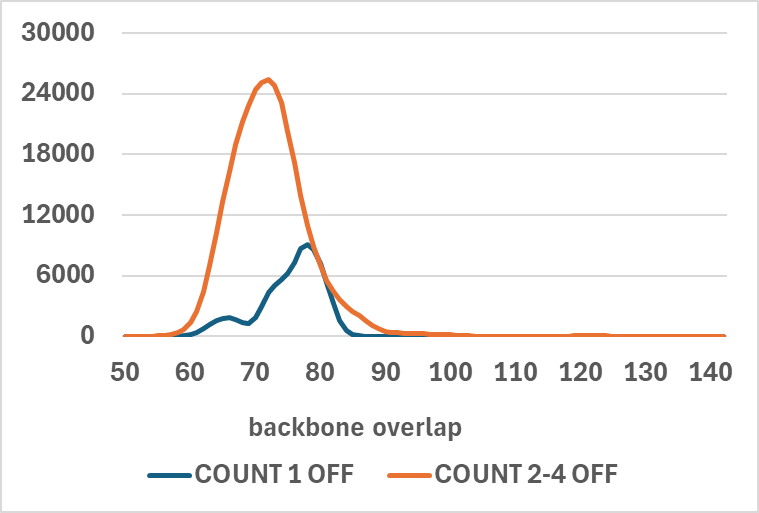}
        \caption{$Seed=3$}
        \label{fig:backbone:hard:file3}
    \end{subfigure}
    \begin{subfigure}[b]{0.33\linewidth}
        \centering
        \includegraphics[scale=0.45]{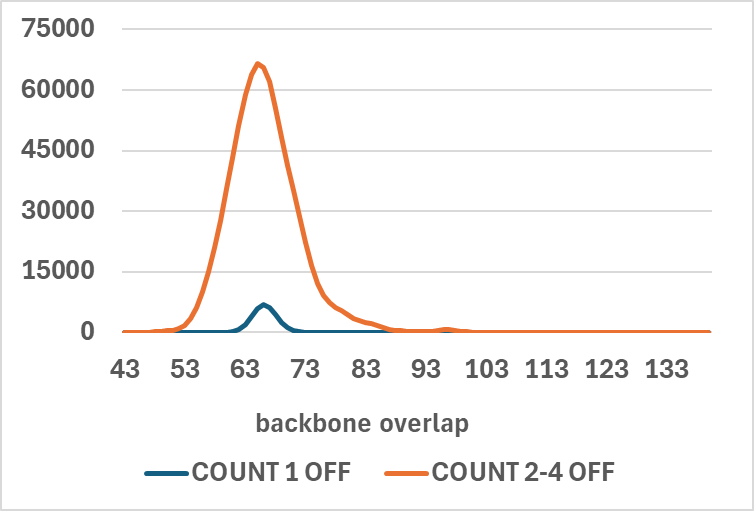}
        \caption{$Seed=14$}
        \label{fig:backbone:hard:file14}
    \end{subfigure}


    \begin{subfigure}[b]{0.33\linewidth}            
        \centering
        \includegraphics[scale=0.45]{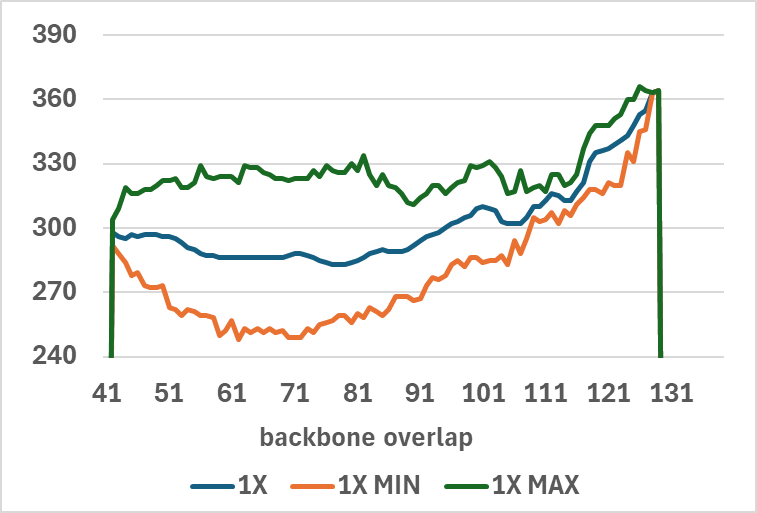}
        \caption{$Seed=2$, 1-time satisfied clauses}
        \label{backbone:hard:file_2_1x}
    \end{subfigure}
    \begin{subfigure}[b]{0.33\linewidth}            
        \centering
        \includegraphics[scale=0.45]{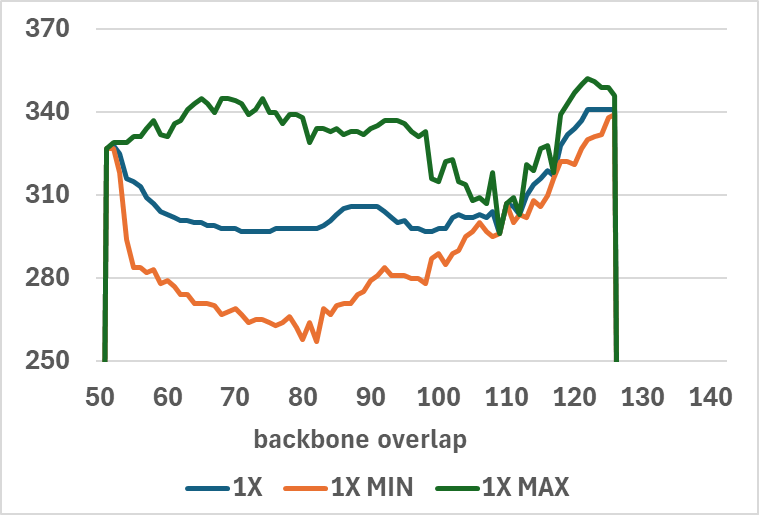}
        \caption{$Seed=3$, 1-time satisfied clauses}
        \label{backbone:hard:file_3_1x}
    \end{subfigure}
    \begin{subfigure}[b]{0.33\linewidth}            
        \centering
        \includegraphics[scale=0.45]{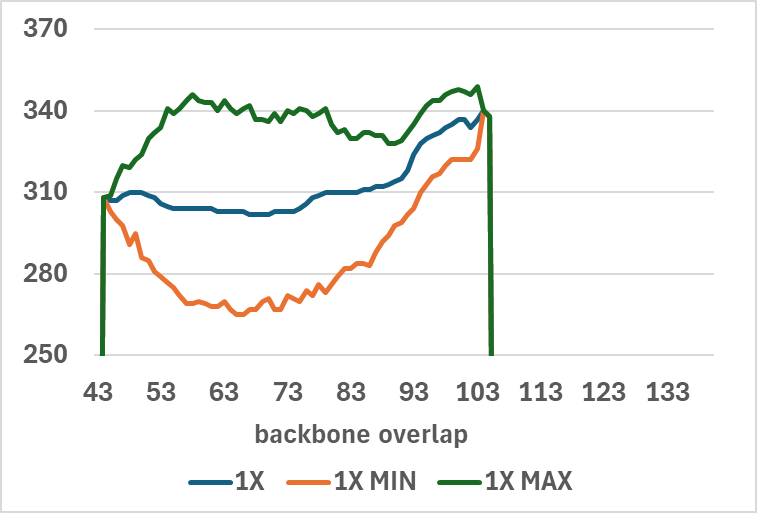}
        \caption{$Seed=14$, 1-time satisfied clauses }
        \label{backbone:hard:file_14_1x}
    \end{subfigure}


    \begin{subfigure}[b]{0.33\linewidth}            
        \centering
        \includegraphics[scale=0.45]{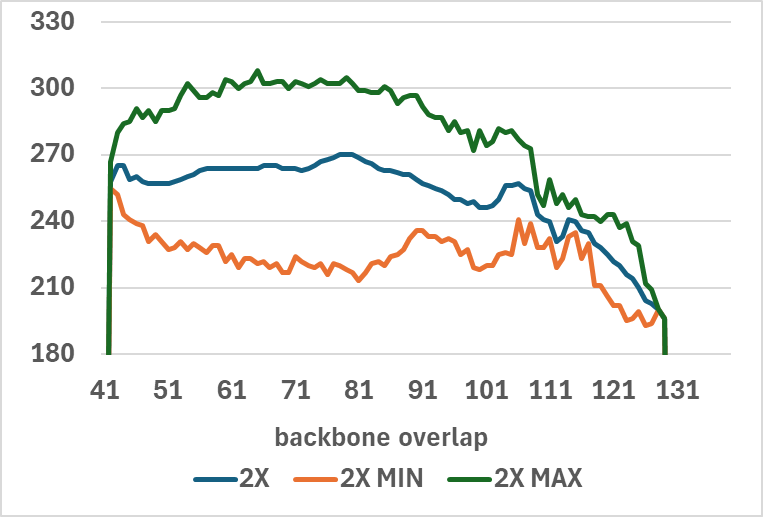}
        \caption{$Seed=2$, 2-times satisfied clauses }
        \label{backbone:hard:file_2_2x}
    \end{subfigure}
    \begin{subfigure}[b]{0.33\linewidth}            
        \centering
        \includegraphics[scale=0.45]{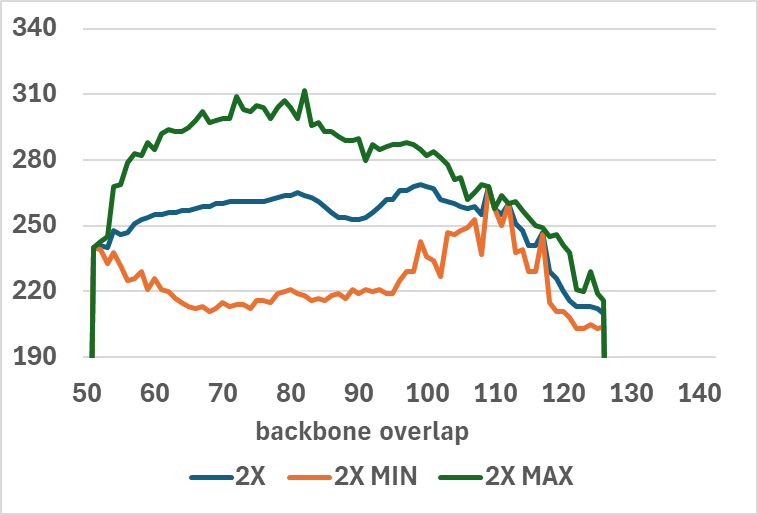}
        \caption{$Seed=3$, 2-times satisfied clauses}
        \label{backbone:hard:file_3_2x}
    \end{subfigure}
    \begin{subfigure}[b]{0.33\linewidth}            
        \centering
        \includegraphics[scale=0.44]{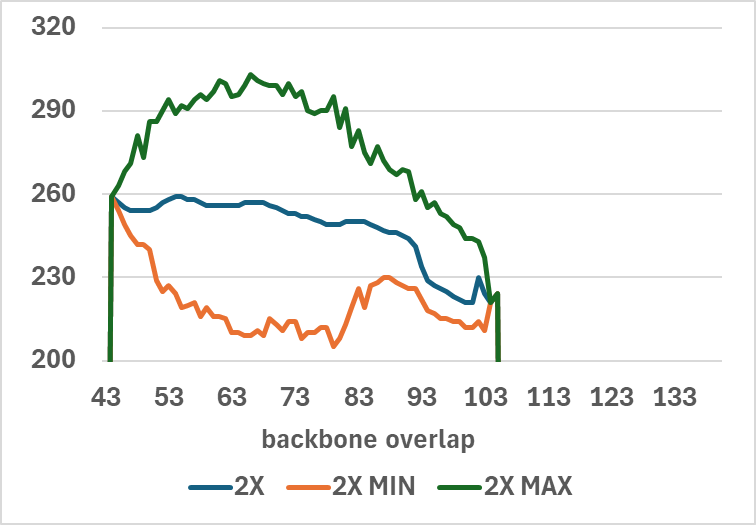}
        \caption{$Seed=14$, 2-times satisfied clauses}
        \label{backbone:hard:file_14_2x}
    \end{subfigure}

    \caption{Backbone overlap distribution for the hard-to-solve instances (X axis: backbone overlap, 
    Y axis: number of solutions for a-c, 1-time satisfied clauses for d-e, 2-times satisfied clauses for g-i)}
    \label{fig:backbone:hard:files}
\end{figure*}

\begin{figure*}[h]


    \begin{subfigure}[b]{0.33\linewidth}            
        \centering
        \includegraphics[scale=0.45]{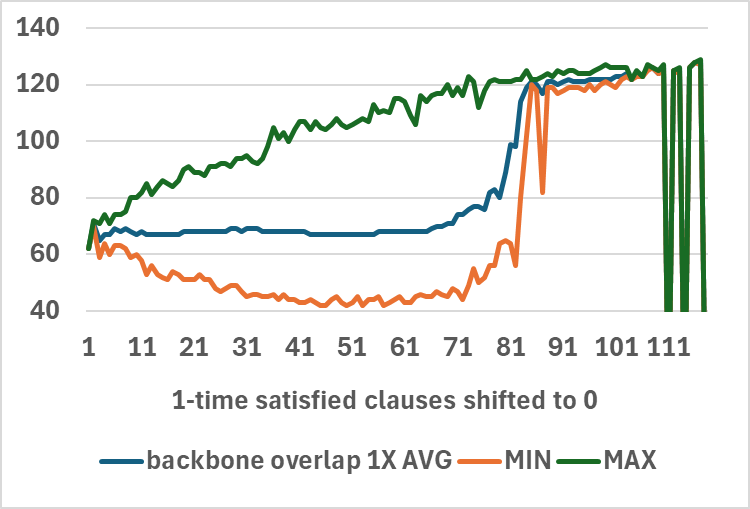}
        \caption{$Seed=2$}
        \label{backbone:hard:file_2_1xrev}
    \end{subfigure}
    \begin{subfigure}[b]{0.33\linewidth}            
        \centering
        \includegraphics[scale=0.44]{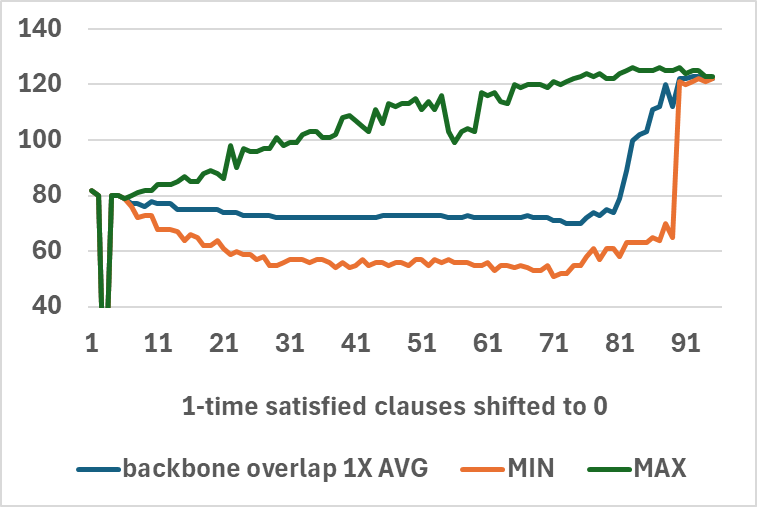}
        \caption{$Seed=3$}
        \label{backbone:hard:file_3_1xrev}
    \end{subfigure}
    \begin{subfigure}[b]{0.33\linewidth}            
        \centering
        \includegraphics[scale=0.45]{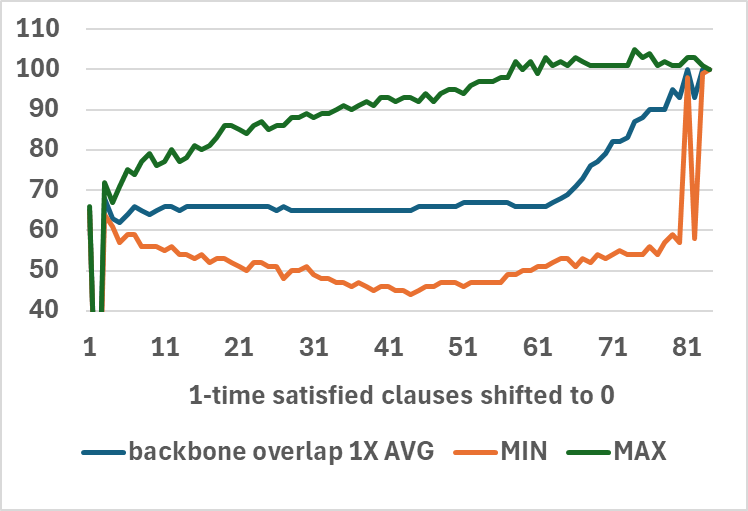}
        \caption{$Seed=14$}
        \label{backbone:hard:file_14_1xrev}
    \end{subfigure}


    \begin{subfigure}[b]{0.33\linewidth}            
        \centering
        \includegraphics[scale=0.45]{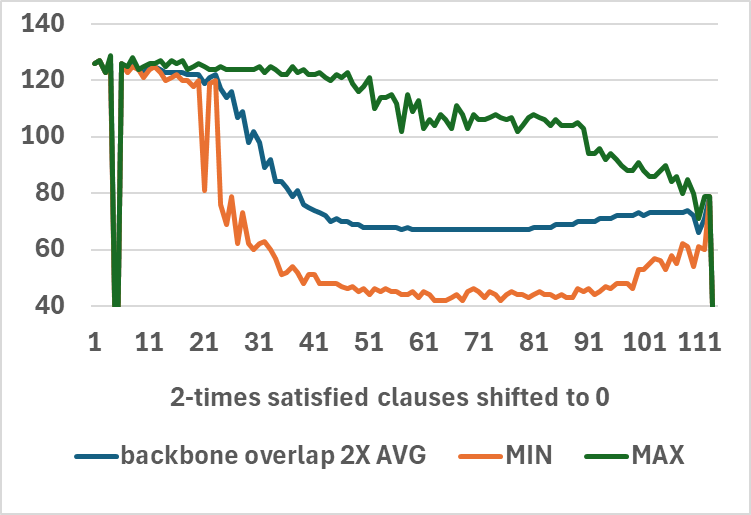}
        \caption{$Seed=2$}
        \label{backbone:hard:file_2_2xrev}
    \end{subfigure}
    \begin{subfigure}[b]{0.33\linewidth}            
        \centering
        \includegraphics[scale=0.45]{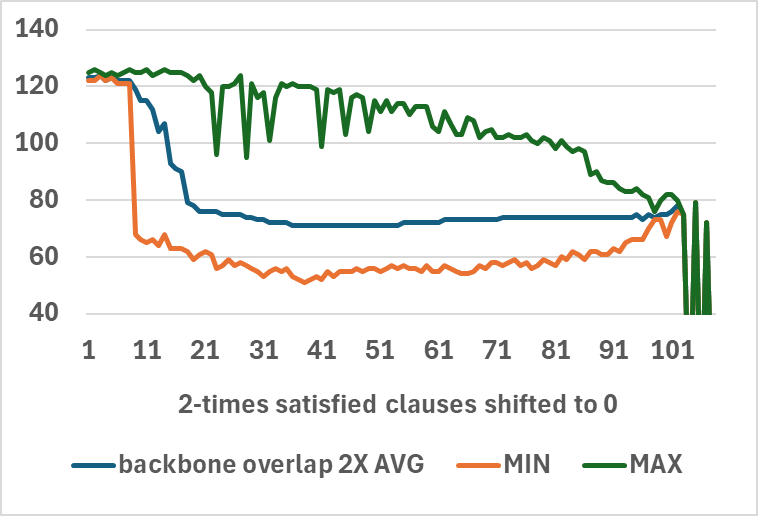}
        \caption{$Seed=3$}
        \label{backbone:hard:file_3_2xrev}
    \end{subfigure}
    \begin{subfigure}[b]{0.33\linewidth}            
        \centering
        \includegraphics[scale=0.45]{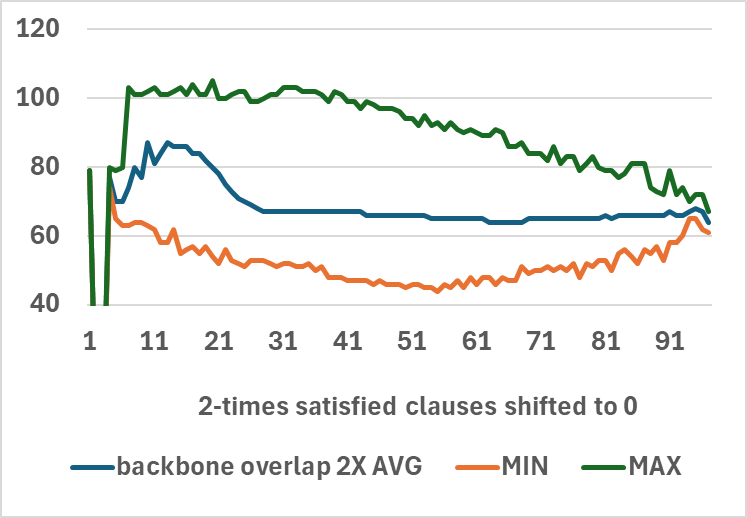}
        \caption{$Seed=14$}
        \label{backbone:hard:file_14_2xrev}
    \end{subfigure}
    
    \caption{Reverted backbone overlap distribution for the hard-to-solve instances (X axis: Number of 1x clauses shifted to 0 for a-c and Number of 2x clauses shifted to 0 for d-f, Y axis: Backbone overlap)}
    \label{fig:backbone:hard:filesReverted}
\end{figure*}

The distribution of high-quality solutions for hard-to-solve instances differs significantly from that of easy-to-solve ones. In Fig.\ref{fig:backbone:hard:file2}-\ref{fig:backbone:hard:file14}, there is no continuous string of solutions from the middle of the distribution to the values that refer to the fully covered backbone. Such backbone overlap distribution indicates that the region of globally optimal solutions must be either found by luck or the optimizer will have to make a large jump from a high-quality solution inside the distribution to the region of globally optimal solutions.The experimental study presented in the latter sections shows that, frequently, tunnelling based on PX and ILS will not be enough to make so large improving moves.\par

As stated at the beginning of this section, the knowledge about backbone is unavailable during the optimization process. Therefore, we wish to find a statistic that can be computed without knowing the optimal solutions and that is correlated with covering the backbone for the hard-to-solve test cases. To this end, we count the number of clauses fulfilled once or twice in a given solution. For instance, solution $100$ fulfils clause $(x_1 \lor \neg x_2 \lor x_3)$ two times because the clause is fulfilled by the first and second variables. By $C_1(\vec{x})$, $C_2(\vec{x})$, and $C_3(\vec{x})$, we denote the number of clauses fulfilled once, twice or three times, respectively. In Fig. \ref{backbone:hard:file_2_1x}-\ref{backbone:hard:file_14_1x}, we show the minimal, average, and maximal value of $C_1(\vec{x})$ with the increase of the backbone coverage. In Fig. \ref{backbone:hard:file_2_2x}-\ref{backbone:hard:file_14_2x}, we show the same statistic but for $C_2(\vec{x})$. These figures suggest that for the hard-to-solve-test cases when we consider high-quality solutions, $C_1(\vec{x})$ increases with the increase of the backbone coverage. At the same time, the value of $C_2(\vec{x})$ decreases with the increase of backbone coverage.\par

In Fig. \ref{fig:backbone:hard:filesReverted}, we show the backbone overlap in the Y-axis (shifted to zero), while the X-axis refers to the number of clauses satisfied by one or two variables. These figures show that maximizing the number of clauses satisfied by one variable is a must to achieve high backbone coverage. Such observation is intuitive. If $C_2(\vec{x})$ is high, then we are free to flip many variables without deteriorating the fitness value because, in the worst case, we will turn some $C_2$ clauses into $C_1$ clauses, but they will remain satisfied. Thus, we can search for improving moves with a low risk of deteriorating an optimized solution. However, if $C_1(\vec{x})$  is high and $C_2(\vec{x})$ is low (we ignore $C_3(\vec{x})$ that is low for high-quality solutions to hard-to-solve instances), then modifying a single variable may turn many $C_1$ clauses into unsatisfied ones and deteriorate solution's fitness. Thus, finding improving moves will be challenging in this case.


\section{Max3Sat Optimizer With Clause-Satisfiability Manipulation}
\label{sec:wmo}
Considering the observations from the previous section, we propose the clause-satisfiability function. Max3Sat Optimizer With Clause-Satisfiability Manipulation (MOCSM) uses this function in the long connection creation step and in the rest of its procedure. The long connection creation step allows to move from solution space regions populated by high-quality but suboptimal solutions close to the region with globally optimal solutions. The long connection mechanism should be useful for those Max3Sat instances for which the tunnelling mechanisms are unable to perform moves that are long enough. \par

We propose the following clause-satisfiability function to rate if the modified version ($x_{mod}$) of a given solution $\vec{x}$ is a step towards the region of globally optimal solutions.
\begin{equation}
    \label{eq:altFitness}
    f_{cf}(\vec{x_{mod}}, \vec{x}) = (C_1(\vec{x_{mod}}) - C_1(\vec{x})) - (C_2(\vec{x_{mod}}) - C_2(\vec{x}))
\end{equation}

The proposed $f_{cf}$ arises from the intuitions described in the prior section. For hard-to-solve instances, increasing $(C_1(\vec{x_{mod}}) - C_1(\vec{x}))$ and decreasing $(C_2(\vec{x_{mod}}) - C_2(\vec{x}))$ shall lead to the regions with globally optimal solution.\par

The rest of this section is organized as follows. In the first subsection, we describe the Max3Sat multi-satisfiability table (MMST), which is a gray-box-like structure utilized by MOCSM and the \textit{ILS+P3+PX} optimizer (IPP), which we use as a baseline. Defining and using MMST is purely technical, but its detailed description shall clarify why some steps in MOCSM and IPP are easy to identify. The second subsection describes IPP, while the third subsection presents MOCSM, including the long connection creation mechanism.

\subsection{Max3Sat multi-satisfiability table}
\label{sec:wmo:mmst}
The Max3Sat Multi-Satisfiability Table (MMST) is a gray-box-like structure that significantly improves the efficiency of finding the potential improving moves. MMST is created for a given solution, and it stores the results of $C_{cf}(\vec{x},x_i)$, i.e., the number of clauses taking $x_i$ as an argument and are satisfied $cf$ times. For instance, $C_{0}(\vec{x},x_5)$ is the number of unsatisfied clauses taking $x_5$ as an argument. Additionally, MMST stores the values of $S_{cf}(\vec{x},x_i)$ and $U_{cf}(\vec{x},x_i)$, which consider the satisfied clauses taking $x_i$ as an argument, for which $x_i$ is among the arguments that satisfy the clause or is not among such arguments, respectively.\par

$C_{0}(\vec{x},x_i)$ informs how many clauses will be satisfied by flipping $x_i$. $S_{1}(\vec{x},x_i)$ is the opposite information, i.e., how many clauses will become unsatisfied after flipping $x_i$. Thus, $C_{0}(\vec{x},x_i) - S_{1}(\vec{x},x_i)$ is the fitness change after flipping $x_i$. Note that flipping $x_i$ will cause $C_{0}(\vec{x},x_i)$ and $S_{1}(\vec{x},x_i)$ to swap their values. Similarly, flipping $x_i$ will swap the following value pairs: ($S_{2}(\vec{x},x_i)$ and $U_{1}(\vec{x},x_i)$) and ($S_{3}(\vec{x},x_i)$ and $U_{2}(\vec{x},x_i)$).\par

Experimental verification shows that, for the considered source code, the cost of computing MMST from scratch is approximately 2-5 times higher than the cost of computing regular fitness. However, using MMST, fitness changes caused by flipping a single variable can be recomputed at a fraction of the regular cost. Thus, MMST makes using FIHC (and other local search algorithms based on modifying a single variable) extremely efficient. For instance, it shows which bit-flip moves will improve fitness. The same or similar concepts were already reported in similar research \cite{mmts}. An example of the MMST use can be found in Section S-I (supplementary material).\par

\subsection{ILS+P3+PX Optimizer}
\label{sec:wmo:ipp}

\begin{algorithm}[h]
	\caption{ILS+P3+PX Optimizer}
	\begin{algorithmic}[1]
            \State $P3  \leftarrow empty$; \label{line:ipp:p3Init}
            \While {$\neg StopCondition$}
                \For {$level = 0$ to size($P3$)-1} \label{line:ipp:ilsStart}
                    \For{\textbf{each} $solution$ \textbf{in} $P3[level]$}
                        \State $solution  \leftarrow ILS(solution)$; \label{line:ipp:ilsEnd}
                    \EndFor 
                \EndFor
                \State $climber \leftarrow randomSolution()$; \label{line:ipp:climber}
                \State $climber  \leftarrow ILS(climber)$;\label{line:ipp:climberILS}
                \State $P3[0] \leftarrow P3[0] + climber$; \label{line:ipp:climbAdd}
                
                \State $climberMod  \leftarrow empty$;
                \For {$level = 0$ to size($P3$)-1} \label{line:ipp:PXstart}
                    \For{\textbf{each} $solution$ \textbf{in} $P3[level]$}
                        \State $climberMod  \leftarrow PX(climber, solution)$;
                        \If {Fit($climberMod$) $>$ Fit($climber$)}
                            \State $P3[level+1] \leftarrow P3[level+1] + climberMod$;
                            \State $climber  \leftarrow climberMod$; \label{line:ipp:PXend}
                        \EndIf
                    \EndFor 
                \EndFor
                
            \EndWhile
	\end{algorithmic}
	\label{alg:ipp}
\end{algorithm}

LS+P3+PX Optimizer (IPP) joins PX and dependency-aware ILS procedure with P3-like population management. The pyramid of solutions is initialized empty (line \ref{line:ipp:p3Init}). At the beginning of each iteration, every individual in the pyramid is optimized with the ILS-like procedure (lines \ref{line:ipp:ilsStart}-\ref{line:ipp:ilsEnd}). The perturbation mask is constructed as follows. We randomly choose a clause (subfunction). Then, we find all clauses that share at least one argument with the chosen clause. We join the arguments of all these subfunctions in one mask and reduce its size by randomly choosing 25\% of its variables. Finally, we randomly re-initialize the variables covered by the mask and execute FIHC to optimize the individual. If the resulting individual is of the same or higher quality, it is preserved or rejected otherwise. \par

After improving every individual with the ILS-like procedure, in each iteration a new individual ($climber$) is created (line \ref{line:ipp:climber}), optimized by the ILS-like procedure (line \ref{line:ipp:climberILS}), and added to the first level of the pyramid (line \ref{line:ipp:climbAdd}). Then, using PX, $climber$ is mixed with the individuals on subsequent pyramid levels (lines \ref{line:ipp:PXstart}-\ref{line:ipp:PXend}). If any of these operations improves $climber$, then its improved copy is added one level higher than the individual the $climber$ is mixed with.\par

\subsection{Proposed Optimizer}
\label{sec:wmo:wmo}

\begin{algorithm}[h]
	\caption{DirectedFIHC function}
	\begin{algorithmic}[1]
    
        \Function{DirectedFIHC}{$solution$}
            \State $availableImpr \leftarrow getAvailableImpr(solution)$; \label{line:dirFIHC:avMoves}
            \While {$availableImpr.size() > 0$}
                \State $clauseFillList \leftarrow empty$;
                \For{\textbf{each} $bitMod$ \textbf{in} $availableImpr$}
                    \State $val  \leftarrow  f_{cf}(solution + bitMod, solution)$;
                    \State $clauseFillList  \leftarrow clauseFillList + (val,bitMod)$;
                \EndFor 
                \State $bitToChange \leftarrow maxElement(clauseFillList)$;
                \State $solution[bitToChange] \leftarrow \neg solution[bitToChange]$;
                \State $availableImpr \leftarrow getAvailableImpr(solution)$;
            \EndWhile

            \State \Return{$solution$}
        \EndFunction 
	\end{algorithmic}
	\label{alg:dirFIHC}
\end{algorithm}

\begin{algorithm}[h]
	\caption{DirectedILS function}
	\begin{algorithmic}[1]
    
        \Function{DirectedILS}{$solution$}
            \State $oldValue \leftarrow Fit(solution)$;
            \State $solOld \leftarrow solution$;
            \State $unsatClauses \leftarrow getUnsatisfiedClauses(solution)$;
            \State $randUnsClause \gets $ getRandClause($unsatClauses$);
            \State $pertMask \leftarrow createMask(randUnsClause)$;
            \State $solution \leftarrow randomize(solution,pertMask)$;
            \State $solution \leftarrow DirectedFIHC(solution)$;
            \If {Fit($solution$) $<$ $ oldValue$}
                \State $solution \leftarrow solOld$;
            \EndIf
            \State \Return{$solution$}
        \EndFunction 
	\end{algorithmic}
	\label{alg:dirILS}
\end{algorithm}

\begin{algorithm}[h]
	\caption{LongConnection function}
	\begin{algorithmic}[1]
    
        \Function{LongConnection}{$solution$, $stepsLimit$}
            \State $steps \leftarrow 0$
            \While {$steps < stepsLimit$} \label{line:warmhole:stop}
            
                \State $clauseFillList \leftarrow empty$; \label{line:warmhole:clauseListStart}
                \For{\textbf{each} $x_i$ \textbf{in} $solution$}
                    \State $val  \leftarrow  f_{cf}(solution + mod(x_i), solution)$;
                    \State $clauseFillList  \leftarrow clauseFillList + (val,mod(x_i))$; \label{line:warmhole:clauseListEnd}
                \EndFor 
                \If {$maxVal(clauseFillList) < 0$}
                    \State \textbf{break}; \label{line:warmhole:broken}
                \EndIf
                \State $bitToChange \leftarrow maxElement(clauseFillList)$; \label{line:warmhole:moveChoose}
                \State $solution[bitToChange] \leftarrow \neg solution[bitToChange]$;  
                \State $steps \leftarrow steps+1$
            \EndWhile

            \State \Return{$solution$}
        \EndFunction 
	\end{algorithmic}
	\label{alg:warmhole}
\end{algorithm}

\begin{algorithm}[h]
	\caption{Max3Sat Optimizer With Clause-Satisfiability Manipulation}
	\begin{algorithmic}[1]
    
        \State $P3  \leftarrow empty$; \label{line:wmo:p3Init}
            \While {$\neg StopCondition$}
                \For {$level = 0$ to size($P3$)-1}
                    \For{\textbf{each} $solution$ \textbf{in} $P3[level]$}
                        \State $solution  \leftarrow DirectedILS(solution)$; \label{line:wmo:directILS}
                    \EndFor 
                \EndFor
                \State $climber \leftarrow randomSolution()$;
                \State $climber  \leftarrow DirectedILS(newSolution)$;
                \State $P3[0] \leftarrow P3[0] + climber$; \label{line:wmo:climbAdd}
                \State $climberMod  \leftarrow empty$;
                \For {$level = 0$ to size($P3$)-1}
                    \For{\textbf{each} $solution$ \textbf{in} $P3[level]$}
                        \State $climberMod  \leftarrow PX(climber, solution)$;
                        \If {Fit($climberMod$) $>$ Fit($climber$)}
                            \State $P3[level+1] \leftarrow AddToP3(climberMod)$;
                            \State $climber  \leftarrow climberMod$;
                        \EndIf
                    \EndFor 
                \EndFor
                
                \For {$level = 0$ to size($P3$)-1} \label{line:wmo:wormStart}
                    \For{\textbf{each} $solution$ \textbf{in} $P3[level]$}
                        \State $solution  \leftarrow LongConnection(solution,25)$; \label{line:wmo:wormEnd}
                    \EndFor 
                \EndFor
            
            \EndWhile
	\end{algorithmic}
	\label{alg:wmo}
\end{algorithm}

In MOCSM, we employ MMST to leverage the efficiency of the considered mechanisms and incorporate the proposed $f_{cf}$. We describe them first. Then, we present the general MOCSM procedure. \par

In directed FIHC (Pseudocode \ref{alg:dirFIHC}), we improve the processed solution, but we choose those improving moves that yield the highest values of $f_{cf}$. The motivation behind this procedure is to choose those improving moves that tend to shift the optimized solution towards the globally optimal solutions region during the local search optimization. First, using MMST, we get the list of available improving bit flips (line \ref{line:dirFIHC:avMoves}). We choose the best bit flip concerning $f_{cf}$ (if more than one move have the same highest $f_{cf}$ value, then we randomly choose one of them). The procedure is repeated until no improving bit flips are left. \par

Directed ILS is presented in Pseudocode \ref{alg:dirILS}. Using MMST, we obtain a list of unsatisfied clauses, randomly choose one of those, and create a perturbation mask in the same way as for IPP (see Section \ref{sec:wmo:ipp}). Then, as in IPP, we optimize the processed solution (but use directed FIHC instead of FIHC). If this operation improves the optimized individual, its modified version is preserved or rejected otherwise.\par

The proposed LongConnection-using procedure is presented in Pseudocode \ref{alg:warmhole}. The motivation behind this procedure is to use $f_{cf}$ to shift the processed solution towards the region occupied by the globally optimal solutions. The procedure is held for a given number of steps (line \ref{line:warmhole:stop}). In this work, we set this value to $25$. At each iteration, we create a ranking of bit flip moves concerning $f_{cf}$ (lines \ref{line:warmhole:clauseListStart}-\ref{line:warmhole:clauseListEnd}). If there are no available moves related to the positive $f_{cf}$ value, then the procedure is stopped (line \ref{line:warmhole:broken}). We choose a move that is related to the highest $f_{cf}$ value (line \ref{line:warmhole:moveChoose}).\par

The general procedure of MOCSM is based on the IPP procedure and is presented in Pseudocode \ref{alg:wmo}. Instead of the standard dependency-aware ILS, it uses directed ILS. MOCSM extends IPP by processing every individual by the LongConnection-using procedure at the end of each iteration (lines \ref{line:wmo:wormStart}-\ref{line:wmo:wormEnd}).\par

To supplement the above optimizer set, we also propose a MOCSM-mixed optimizer that is supposed to be an intermediate version between MOCSM and IPP. In MOCSM-mixed, at the beginning of each iteration, for every second individual, we use standard ILS and directed ILS for the other half of the population. Similarly, at the end of the iteration, we use the LongConnection procedure for every second individual while the rest of the population remains unaffected.

\section{Max3Sat instance difficulty}
\label{sec:difficulty}

\subsection{Proposed difficulty measurement procedure}
\label{sec:difficulty:measure}
As presented in Section \ref{sec:backbone}, the shape of the backbone overlap distribution seems correlated with the instance difficulty. To verify if our observations are valid we propose the following procedure. We use the results of the IPP runs employed to generate backbone (the detailed description of which solutions we take into account is reported in Section \ref{sec:backbone}). For the 10\% of the best-found solutions, we compute the average of:

\begin{equation}
    \label{eq:meas}
    backDist(\vec{x},\vec{backbone})= |\vec{backbone}| - over(\vec{x}, \vec{backbone})
\end{equation}
where $\vec{backbone}$ is the backbone found for the considered instance, $|\vec{backbone}|$ is the number of variable values specified by backbone, and $over(\vec{x}, \vec{backbone})$ is the number of equal variables in $\vec{x}$ and the variables specified by backbone.\par

The intuition behind the proposed measure is that if the best-found individuals are close to the backbone, then the region with globally optimal solutions should be easy to find (the situation presented in Fig. \ref{fig:backbone:easy:files}). If the value is high, then we deal with the situation presented in Fig. \ref{fig:backbone:hard:file2}-\ref{fig:backbone:hard:file14}.

\subsection{Experiment setup}
\label{sec:difficulty:setup}

We consider two types of Max3Sat instances. The artificial instances form the uniformly random generator employed in \cite{P3Original} and the generator of industrial-like instances used in \cite{francisGen}. For those instances, we were considering 100, 150, and 200 variables and the clause ratio ($cr$) from the set $\{3.0, 3.5, 4.0, 4.27\}$. The industrial-like instance generator creates instances where the degree of the variables in the clause-variable interaction graph follow a power-law distribution with parameter $\beta$, (the so-called scale-free distribution). See \cite{ansotegui} for more details. We considered $\beta$ from the set $\{2,5,8\}$. Industrial-like instances were transformed to Max3Sat \cite{transTokBounded}. As a black-box optimizer, we were using P3, that is parameter-less and was shown effective in solving Max3Sat problem \cite{P3Original,FIHCwLL} with stop condition set on 12 hours. All the code and instances can be found in the replication package accompanying this paper~\cite{zenodo-package} and in GitHub\footnote{\url{https://github.com/plebian1/Max3Sat-multi-satisfiability/tree/main}} \par

\subsection{Difficulty measurement results}
\label{sec:backbone_grading_results}

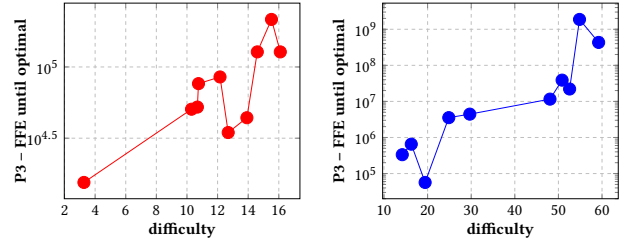
\begin{figure}[h]
    \begin{subfigure}[b]{0.49\linewidth}
		\resizebox{0.95\linewidth}{!}{%
        \tikzset{every mark/.append style={scale=2.5}}
        \begin{tikzpicture}
        \begin{axis}[%
        legend entries={$f_0(u)$, $f_1(u)$, $f_{w=0.5}(u)$, $f_{w=0.3(3)}(u)$, $f_{w=0.6(6)}(u)$},
        legend columns=-1,
        legend to name=named,
        ymode=log,
        xlabel=\textbf{difficulty},
        ylabel=\textbf{P3 -- FFE until optimal},
        grid,
        grid style=dashed,
        ticklabel style={scale=1.5},
        label style={scale=1.5},
        legend style={font=\fontsize{8}{0}\selectfont}
        ]
        
        \addplot[
        color=red,
        mark=*,
        ]
        coordinates {
            (3.26852,15471)(10.297,50512)(10.6806,52286)(10.7525,76405)(12.1595,85088)(12.6943,34660)(13.9159,44021)(14.5956,127600)(15.5217,215647)(16.0973,127580)
        };

        \end{axis}
        \end{tikzpicture}
    }
	\caption{100-bit industrial-like instances $cr=4.0$, $\beta=8$}
	\label{fig:correlation:ind}
	\end{subfigure}
    \begin{subfigure}[b]{0.49\linewidth}
		\resizebox{0.95\linewidth}{!}{%
        \tikzset{every mark/.append style={scale=2.5}}
        \begin{tikzpicture}
        \begin{axis}[%
        legend columns=-1,
        legend to name=named,
        ymode=log,
        xlabel=\textbf{difficulty},
        ylabel=\textbf{P3 -- FFE until optimal},
        grid,
        grid style=dashed,
        ticklabel style={scale=1.5},
        label style={scale=1.5},
        legend style={font=\fontsize{8}{0}\selectfont}
        ]
        
        \addplot[
        color=blue,
        mark=*,
        ]
        coordinates {
            (14.1901,333682)(16.2548,652740)(19.4317,56759)(24.8584,3548805)(29.6797,4438478)(48.0684,11564120)(50.8452,38895782)(52.5701,22079980)(54.8405,1874062362)(59.2222,432076277)

        };

        \end{axis}
        \end{tikzpicture}
    }
	\caption{150-bit artificial instances $cr=4.27$}
	\label{fig:correlation:art}
	\end{subfigure}


	\caption{The correlation between difficulty measurement results and P3 budget necessary to find the optimum}
	\label{fig:correlation}
\end{figure}

\begin{table}  
    \caption{Difficulty measurement results and P3-based FFE budget necessary for finding the optimal solution -- Spearman test results}
    \label{tab:spearman}
    \hspace{1em}
    \begin{subtable}{.22\textwidth}
        \begin{tabular}{c|cc}
            $cr$   & \textbf{Industr}  & \textbf{Artif}  \\
            \hline
            \textbf{3.00} & 0.75 & 0.78 \\
            \textbf{3.50} & 0.58 & 0.75 \\
            \textbf{4.00} & 0.85 & 0.69 \\
            \textbf{4.27} & 0.81 & 0.60
            \end{tabular}
        \caption{Cr-based correlation}
        \label{tab:spearman:cr}
    \end{subtable}
    \begin{subtable}{.22\textwidth}
    \centering
       \begin{tabular}{c|cc}
            \textbf{size}   & \textbf{Industr}  & \textbf{Artif}  \\
            \hline
            \textbf{100}  & 0.72 & 0.79 \\
            \textbf{150}  & 0.65 & 0.49 \\
            \textbf{200}  & 0.50 & 0.46 \\
            \textbf{All}  & 0.69 & 0.68
            \end{tabular}
        \caption{Size-based correlation}
        \label{tab:spearman:size}
    \end{subtable}
\end{table}

In Fig. \ref{fig:correlation}, we show an exemplary correlation for two Max3Sat groups, the industrial-like (Fig. \ref{fig:correlation:ind}) and artificial (Fig. \ref{fig:correlation:art}). For both test case groups, the fitness function evaluation number (FFE) until finding the globally optimal solution by P3 increases (in general) with the increase of the instance difficulty measured using the proposed procedure. The difficulty ranges and FFE budgets differ for both results series. However, some of their parts overlap. Note that although two instance series originate from different generators, they are coherent. In Fig. S-I (supplementary material), we show both curves joined on one figure.\par

The above observations were confirmed by the results of the Spearman test. Its results are reported in Table \ref{tab:spearman}. The correlation depending on $cr$ (Table \ref{tab:spearman:cr}) and size (Table \ref{tab:spearman:size}) can be considered significant. The values of the Spearman test that are above $0.6$ can be considered as related to strong or very strong correlation, while the values above $0.4$ are related to strong or moderate correlation.

\section{Experimental effectiveness comparison}
\label{sec:results}

\begin{table}  
    \caption{Average time [in seconds] for finding the optimal result (200-bit instances)}
    \label{tab:effEasy}
    \scriptsize
    \hspace{-2em}
    \begin{subtable}{.22\textwidth}
        \begin{tabular}{l|cccc}
            $cr=$  & \textbf{3.00} & \textbf{3.50} & \textbf{4.00}  & \textbf{4.27}  \\
            \hline
            \textbf{MOCSM} & 0.37 & 0.91 & 25.64 & 65.85 \\
            \textbf{MIX} & 0.18 & 0.54 & 10.26 & 23.39 \\
            \textbf{IPP} & 0.15 & 0.39 & 12.74 & 17.27
        \end{tabular}
        \caption{Industrial-like}
        \label{tab:effEasy:industr}
    \end{subtable}
    \begin{subtable}{.22\textwidth}
    \centering
       \begin{tabular}{l|cccc}
         $cr=$   & \textbf{3.00} & \textbf{3.50} & \textbf{4.00} & \textbf{4.27}\\
         \hline
        \textbf{MOCSM} & 0.35 & 1.17 & 6.82 & 1.34\\
        \textbf{MIX} & 0.15 & 0.52 & 4.19 & 13.09\\
        \textbf{IPP} & 0.13 & 0.38 & 3.24 & 795.41
        \end{tabular}
        \caption{Articial}
        \label{tab:effEasy:art}
    \end{subtable}
\end{table}

\begin{figure}[h]
    \begin{subfigure}[b]{0.49\linewidth}
		\resizebox{0.95\linewidth}{!}{%
        \tikzset{every mark/.append style={scale=2.5}}
        \begin{tikzpicture}
        \begin{axis}[%
        legend entries={$f_0(u)$, $f_1(u)$, $f_{w=0.5}(u)$, $f_{w=0.3(3)}(u)$, $f_{w=0.6(6)}(u)$},
        legend columns=-1,
        legend entries={P3 (12 hours run), IPP, MOCSM-MIXED, MOCSM},
        legend to name=named,
        xtick={1,2,3,4,5,6},
        xticklabels={100,150,200,250,300,350},
        xlabel=\textbf{size},
        ylabel=\textbf{Solved instance rate},
        grid,
        grid style=dashed,
        ticklabel style={scale=1.5},
        label style={scale=1.5},
        legend style={font=\fontsize{8}{0}\selectfont}
        ]

        \addplot[
        color=black,
        mark=*,
        ]
        coordinates {
            (1,1)(2,1)(3,0.8333333)(4,0.77)
        };

        \addplot[
        color=blue,
        mark=*,
        ]
        coordinates {
            (1,1)(2,0.96666)(3,0.8333333)(4,0.8)(5,0.4)(6,0.5)
        };
        
        \addplot[
        color=violet,
        mark=square*,
        ]
        coordinates {
            (1,1)(2,1)(3,1)(4,1)(5,0.86666)(6,0.8)
        };

        \addplot[
        color=red,
        mark=*,
        ]
        coordinates {
            (1,1)(2,1)(3,1)(4,1)(5,1)(6,1)
        };

        \end{axis}
        \end{tikzpicture}
    }
	\caption{Solved instances rate}
	\label{fig:scalability:rate}
	\end{subfigure}
    \begin{subfigure}[b]{0.49\linewidth}
		\resizebox{0.95\linewidth}{!}{%
        \tikzset{every mark/.append style={scale=2.5}}
        \begin{tikzpicture}
        \begin{axis}[%
        legend entries={P3 (12hours), IPP, MOCSM-MIX, MOCSM},
        legend columns=-1,
        legend to name=named,
        xtick={1,2,3,4,5,6},
        xticklabels={100,150,200,250,300,350},
        xlabel=\textbf{size},
        ylabel=\textbf{P3 -- FFE until optimal},
        grid,
        grid style=dashed,
        ticklabel style={scale=1.5},
        label style={scale=1.5},
        legend style={font=\fontsize{8}{0}\selectfont}
        ]

        \addplot[
        color=black,
        mark=*,
        ]
        coordinates {
            (1,20.5181401)(2,1112.78074949333)
        };

        \addplot[
        color=blue,
        mark=*,
        ]
        coordinates {
            (1,1.17722176333333)(2,231.9400373)
        };
        
        \addplot[
        color=violet,
        mark=square*,
        ]
        coordinates {
            (1,0.354695206666667)(2,2.41016096666667)(3,13.0938504666667)(4,144.627749666667)
        };

        \addplot[
        color=red,
        mark=*,
        ]
        coordinates {
            (1,0.0429108233333333)(2,0.27091688)(3,1.34414471333333)(4,5.9270693)(5,11.8799640666667)(6,42.4123702)
        };

        \end{axis}
        \end{tikzpicture}
    }
	\caption{Time until optimal sol.}
	\label{fig:scalability:time}
	\end{subfigure}

	\ref{named}

	\caption{Scalability analysis for artificial instances $cr=4.27$}
	\label{fig:scalability}
\end{figure}
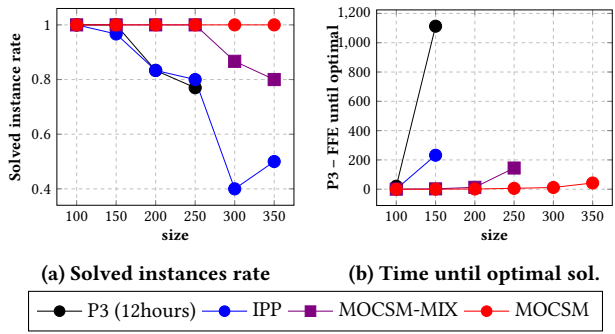

To verify the effectiveness of MOCSM, we consider MOCSM-mixed and IPP as competing optimizers. Each was executed on Intel core i7-14700KF, 64GB RAM with 1 hour of computation time. To show the baseline for the hard-to-solve instances, we report the results of P3 in the black-box setting given 12 hours of computation time. To assure fair comparison, all optimizers were single-threaded, and no other resource-consuming processes were executed.\par

In Table \ref{tab:effEasy}, we report the results for the industrial-like and artificial instances that seem easy to solve for all gray-box optimizers (computation time until finding the optimal solution did not exceed 2 minutes). For all these instances, MOCSM was approximately two times slower than MOCSM-mixed and three times slower than IPP. Such a result is expected because the $f_{cf}$-based mechanisms, including the LongConnection mechanism, decrease the convergence, which is a key element in efficiently solving the easy-to-solve instance. Such a conclusion is confirmed by the results of the MOCSM-mixed between MOCSM and IPP for all considered instance groups.\par

In Fig. \ref{fig:scalability}, we present the scalability analysis for the artificial instances $cr=4.27$ that turned out to be the most challenging to solve. MOCSM solves them all (Fig. \ref{fig:scalability:rate}) with only a little increase of the average computation time necessary for finding the optimal solution (Fig. \ref{fig:scalability:time}). The other optimizers perform significantly worse. P3 and IPP suffer a significant effectiveness drop for 200-bit instances (approximately to 80\%). At the same time the computation time necessary for finding the optimal solution increases significantly. MOCSM-mixed performance is similar to MOCSM at the beginning, but for 250-bit instances the computation time it needs to find the optimal solution increases significantly. For 300-bit instances it suffers a significant decrease in the percentage of successful runs.\par

The above results confirm the intuitions based on the mechanisms proposed in this work. However, they also show that any decrease in the convergence speed results in worse computation times of MOCSM (yet they remain low). However, MOCSM is the only optimizer that solved all considered instances in all runs with computation times that did not exceed 2 minutes. As such, it is the most general, and the increased computation time for solving the easier instances seems to be a reasonable price for the capability of solving those instances that are out of the grasp of all other considered optimizers.\par

\section{Conclusions}
\label{sec:conc}
In this work, we analyze the difficulty of Max3Sat instances, considering concepts arising from phase transitions. This analysis allowed us to identify the features of those instances that seem particularly hard to solve. Using these observations, we proposed the features that can help to identify the regions occupied by globally optimal solutions. On this base, we propose the difficulty-rating mechanism for Max3Sat instances that can be useful for creating benchmark sets. Using a state-of-the-art optimizer, we have confirmed that the proposed difficulty measurement procedure is correct for the considered instances.\par

The proposed LongConnection mechanism is the core of the proposed MOCSM and significantly increases its effectiveness for hard-to-solve instances. Its applicability is limited to Max3Sat, but it is resistant to eventual modifications of the Max3Sat, e.g., adding constraints to the problem definition.\par

The results we present indicate that filtering and analyzing the hard-to-solve problem instances to identify their features have an exceptional potential of proposing highly-effective optimizers. Therefore, the core objective of the future research will be proposing similar mechanisms for other problems also from non-binary domains. Another important research direction is extending the existing Max3Sat benchmark sets by the difficulty indicator proposed in this work. Finally, it is also worth to improve MOCSM by eliminating the sewed-in parameters it uses within its procedure.

\begin{acks}
Jedrzej Piatek and Michal Przewozniczek were supported by the Polish National Science Centre (NCN) under grant 2022/45/B/ST6/04150. Francisco Chicano was supported by the University of M\'alaga under grant PAR 4/2023. Renato Tin\'os was supported by National Council for Scientific and Technological Development under CNPq grant \#306689/2021-9 and by FAPESP under grant \#2024/15430-5.
\end{acks}

	\bibliographystyle{ACM-Reference-Format}
	\bibliography{BlowTunnel} 
	
\end{document}